\documentclass[11pt]{article}
\usepackage{acl2015}
\usepackage{times}
\usepackage{url}
\usepackage{latexsym}
\usepackage{xspace}
\usepackage{booktabs}
\usepackage{color}
\usepackage{graphicx}
\usepackage{tabulary}

\usepackage{amsfonts}

\usepackage{microtype}
\usepackage{multirow}
\usepackage{url}
\usepackage{verbatim}
\usepackage{amsmath,amsthm,amssymb}
\usepackage{array}
\usepackage[scaled=0.86]{helvet}
\usepackage{ifthen}
\usepackage{courier}
\usepackage[linesnumbered,vlined,ruled]{algorithm2e}

\newcommand{\captionfonts}{\small}
\makeatletter  
\long\def\@makecaption#1#2{%
  \vskip\abovecaptionskip
  \sbox\@tempboxa{{\captionfonts #1: #2}}%
  \ifdim \wd\@tempboxa >\hsize
    {\captionfonts #1: #2\par}
  \else
    \hbox to\hsize{\hfil\box\@tempboxa\hfil}%
  \fi
  \vskip\belowcaptionskip}
\makeatother   

\belowdisplayskip \abovedisplayskip

\newcommand{\dbleu}{{$\Delta${\sc Bleu}}\xspace}
\newcommand{\bleu}{{{\sc Bleu}}\xspace}
\newcommand{\meteor}{{{\sc Meteor}}\xspace}
\newcommand{\camready}[1]{{{#1}}}
\title{\dbleu: A Discriminative Metric for Generation Tasks\\ with Intrinsically Diverse Targets\thanks{$\,\,\,$To appear at ACL in July 2015 (submitted April 30, 2015, accepted June 9, 2015).}}

\author{Michel Galley\hspace{1pt}$^{{\bf 1}\dag}$ \hspace{.5cm}
Chris Brockett\hspace{1pt}$^{{\bf 1}}$ \hspace{.5cm}
Alessandro Sordoni\hspace{1pt}$^{{{\bf 2}\ddag}}$ \hspace{.5cm}
Yangfeng Ji\hspace{1pt}$^{{\bf 3}\ddag}$\\[0.05cm]
{\bf Michael Auli\hspace{1pt}$^{{\bf 4}\ddag}$ \hspace{.43cm}
Chris Quirk\hspace{1pt}$^{{\bf 1}}$ \hspace{.43cm}
Margaret Mitchell\hspace{1pt}$^{\bf 1}$  \hspace{.43cm}
Jianfeng Gao\hspace{1pt}$^{\bf 1}$ \hspace{.43cm}
Bill Dolan\hspace{1pt}$^{\bf 1}$}
\\[0.3cm]
{$^1$Microsoft Research, Redmond, WA, USA} \\
{$^2$DIRO, Universit\'e de Montr\'eal, Montr\'eal, QC, Canada} \\
{$^3$Georgia Institute of Technology, Atlanta, GA, USA}\\
{$^4$Facebook AI Research, Menlo Park, CA, USA}\\
}

\date{}

\begin{document}
\maketitle

{\let\thefootnote\relax\footnotetext{\camready{$^\dag$Corresponding author:
{\textsf{mgalley@microsoft.com}}}}}
{\let\thefootnote\relax\footnotetext{\camready{$\ddag$The entirety of this work was conducted while at Microsoft Research.}}}

\begin{abstract}
We introduce Discriminative \bleu (\dbleu), a novel metric for intrinsic evaluation of generated text in tasks that admit a diverse range of possible outputs.  
Reference strings are scored for quality by human raters on a scale of [$-1$, $+1$] to weight multi-reference \bleu. 
In tasks involving generation of conversational responses, \dbleu correlates reasonably with human judgments and outperforms sentence-level and IBM \bleu in terms of both Spearman's $\rho$ and Kendall's $\tau$. 
\end{abstract}

\section{Introduction}
\label{sec:intro}
Many natural language processing tasks involve the generation of texts where a variety of outputs are acceptable or even desirable.
Tasks with intrinsically diverse targets range from machine translation, summarization, sentence compression, paraphrase generation, and image-to-text to generation of conversational interactions.
A major hurdle for these tasks is automation of evaluation, since the space of plausible outputs can be enormous, and it is it impractical to run a new human evaluation every time a new model is built or parameters are modified.

In Statistical Machine Translation (SMT), the automation problem has to a large extent been ameliorated by metrics such as \bleu~\cite{BLEU} and \meteor~\cite{METEOR}
Although \bleu is not immune from criticism (e.g., Callison-Burch et al.~\shortcite{Callison-Burch2006}), its properties are well understood, \bleu scores have been shown to correlate well with human judgments~\cite{Doddington:2002,Coughlin,Graham2014,Graham2015} in SMT, and it has allowed the field to proceed.

\bleu has been less successfully applied to non-SMT generation tasks owing to the larger space of plausible outputs. As a result, attempts have been made to adapt the metric. 
To foster diversity in paraphrase generation, Sun and Zhou ~\shortcite{DBLP:conf/acl/SunZ12} propose a metric called i\bleu in which the \bleu score is discounted by a \bleu score computed between the source and paraphrase. 
This solution, in addition to being dependent on a tunable parameter, is specific only to paraphrase. In image captioning tasks, Vendantam et al.~\shortcite{VedantamZP14}, employ a variant of \bleu in which n-grams are weighted by \emph{tf}{$\cdot$}\emph{idf}.  
This assumes the availability of a corpus with which to compute \emph{tf}{$\cdot$}\emph{idf}. 
Both the above can be seen as attempting to capture a notion of target goodness that is not being captured in \bleu.

In this paper, we introduce Discriminative \bleu (\dbleu), a new metric that embeds human judgments concerning the quality of reference sentences directly into the computation of corpus-level multiple-reference \bleu.
In effect, we push part of the burden of human evaluation into the automated metric, where it can be repeatedly utilized. 

Our testbed for this metric is data-driven conversation, a field that has begun to attract interest~\cite{Ritter2011,Sordoni2015} as an alternative to conventional rule-driven or scripted dialog systems.
Intrinsic evaluation in this field is exceptionally challenging because the semantic space of possible responses resists definition and is only weakly constrained by conversational inputs.

Below, we describe \dbleu and investigate its characteristics in comparison to standard \bleu in the context of conversational response generation.
We demonstrate that \dbleu correlates well with human evaluation scores in this task and thus can provide a basis for automated training and evaluation of data-driven conversation systems---and, we ultimately believe, other text generation tasks with inherently diverse targets.

%

\section{Evaluating Conversational Responses}
\label{sec:metric}
\begin{figure}[tbp]
\centering
\includegraphics[width=7cm]{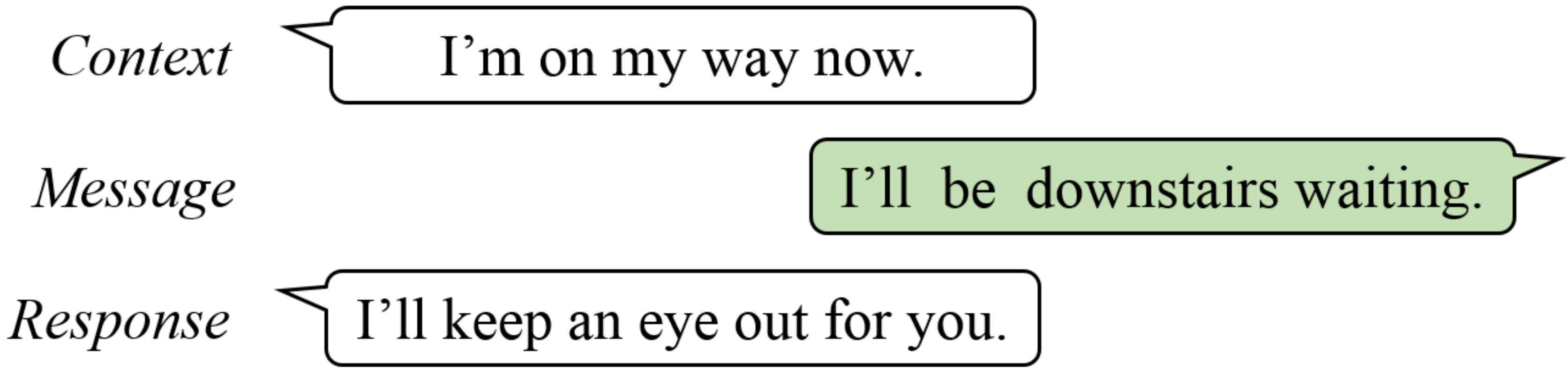}
\caption{Example of consecutive utterances of a dialog.}
\label{tab:sample} 
\end{figure}

Given an input message $m$ and a prior conversation history $c$, the goal of a response generation system is to produce a hypothesis $h$ that is both well-formed and a pertinent response to message $m$ (example in Fig.~\ref{tab:sample}).
We assume that a set of $J$ references $\{r_{i,j}\}$ is available for the context $c_i$ and message $m_i$, where $i \in \{1\ldots{}I\}$ is an index over the test set.
In the case of \bleu,\footnote{\camready{Unless mentioned otherwise, \bleu refers to the original IBM \bleu as first described in \cite{BLEU}.}}
the automatic score of the system output $h_1\ldots{}h_I$ is defined as:\\[-0.2cm]
\begin{equation}
\textrm{\bleu} = \textrm{{\sc BP}} \cdot \exp \bigg( \sum_n \log p_n \bigg)
\label{eq:bleu}
\end{equation}\\[-0.5cm]
with:\\[-0.5cm]
\begin{equation}
\textrm{{\sc BP}} =
\begin{cases}
\, 1           & \textrm{if} \, \, \, \eta > \rho\\
\, 
\, e^{(1-\rho/\eta)} & \textrm{otherwise}
\label{eq:bp}
\end{cases}
\end{equation}
where $\rho$ and $\eta$ are respectively hypothesis and reference lengths.\footnote{In the case of multiple references, \bleu selects the reference whose length is closest to that of the hypothesis.} Then corpus-level $n$-gram precision is defined as:
\begin{equation*}
p_n = \frac{\sum_i \sum_{g \, \in \, n\textrm{-grams}(h_i)}\,\,\max_j\big\{\#_g(h_i, r_{i,j})\big\}}{\sum_i \sum_{g \, \in \, n\textrm{-grams}(h_i)}\,\,\#_g(h_i)}
\end{equation*}
where $\#_g(\cdot)$ is the number of occurrences of \mbox{$n$-gram} $g$ in a given sentence, and $\#_g(u, v)$ is a shorthand for \mbox{$\min\big\{\#_g(u), \#_g(v)\big\}$}.

It has been demonstrated that metrics such as \bleu show increased correlation with human judgment as the number of references increases \cite{MetricsMATR,Dreyer2012}.
Unfortunately, gathering multiple references is difficult in the case of conversations.
Data gathered from naturally occurring conversations offer only one response per message.
\camready{One could search $(c,m)$ pairs that occur multiple times in conversational data with the hope of finding distinct responses, but this solution is not feasible. Indeed, the larger the context, the less likely we are to find pairs that match exactly}.
Furthermore, while it is feasible to have writers create additional references when the downstream task is relatively unambiguous (e.g., MT), this approach is more questionable in the case of more subjective tasks such as conversational response generation.
Our solution is to mine candidate responses from conversational data and have judges rate the quality of these responses. 
Our new metric thus naturally incorporates qualitative weights associated with references.

\section{Discriminative \bleu}
\label{sec:dbleu}

Discriminative \bleu, or \dbleu, extends \bleu by exploiting human qualitative judgments $w_{i,j} \in [-1,+1]$ associated with references $r_{i,j}$.
It is discriminative in that it both rewards matches with ``good'' reference responses ($w > 0$) and penalizes matches with ``bad'' reference responses ($w < 0$). 
Formally, \dbleu is defined as in Equation~\ref{eq:bleu} and~\ref{eq:bp}, except that $p_n$ is instead defined as:
%
\begin{equation*}
\!
\!
\frac{\sum_i \! \sum_{g \, \in \, n\textrm{-grams}(h_i)}\max_{j : g \, \in \, r_{i,j}}\!\big\{w_{i,j} \! \cdot \! \#_g(h_i, r_{i,j})\big\}}{\sum_i \! \sum_{g \, \in \, n\textrm{-grams}(h_i)}\max_j \!\big\{ w_{i,j} \! \cdot \! \#_g(h_i)\big\}}
\end{equation*}
In a nutshell, this is saying that each $n$-gram match is weighted by the highest scoring reference in which it occurs, and this weight can sometimes be negative.
To ensure that the denominator is never zero, we assume that, for each $i$ there exists at least one reference $r_{i,j}$ whose weight $w_{i,j}$ is strictly positive.
In addition to its discriminative nature, this metric has two interesting properties.
First, if all weights $w_{i,j}$ are equal to 1, then the metric score is identical to \bleu.
As such, \dbleu admits \bleu as a special case.
Second, as with IBM \bleu, the maximum theoretical score is also 1.
If the hypothesis happens to match the highest weighted reference for each sentence, the numerator equals the denominator and the metric score becomes 1.
While we find this metric particularly appropriate for response generation, the metric makes no assumption on the task and is applicable to other text generation tasks such as MT and image captioning.
 
%

\section{Data}
\label{sec:data}
\subsection{Multi-reference Datasets}
\label{sec:multiref}

\begin{table*}[tpb]
\centering
\scriptsize
\renewcommand{\arraystretch}{1.4}
\begin{tabular}{p{4.6cm}p{4.5cm}p{4.4cm}r}
\toprule
{\bf Context $c$} & {\bf Message $m$} & {\bf Response $r$} & {\bf Score}\\ \cline{1-3}
\multicolumn{1}{|p{4.6cm}}{
i was about to text you and my two cousins got excited cause they thought you were ``rihanna''}&
aww, i can imagine their disappointment &
\multicolumn{1}{p{4.6cm}|}{
they were very disappointed!!!}                 										       				& 0.6\\ \cline{1-2}
{\it yes. my ex-boyfriend, killed my cat. like i say, it was the start of a bad time...}&
{\it i can imagine!}&
\multicolumn{1}{|p{4.6cm}|}{
{\it yes. luckily, the whole thing feels very much of the past now.}}      				& 0.8\\
{\it its good.. for some reason i can't name stand out tracks but i've been playing it since it dropped}&
{\it i can imagine, banks doesn't disappoint}&
\multicolumn{1}{|p{4.6cm}|}{
{\it na this is anything but a disappointment..}}                          				& 0.6\\
{\it at my lil cousins dancing to ``dance for you''. these kids are a mess.}&
{\it lmaoo i can imagine.}&
\multicolumn{1}{|p{4.6cm}|}{
{\it they were belly rolling, filarious.}}               									 				& 0.4\\
{\it what's sick about it?? do you know how long it is?? no so how is it sick?}&
{\it i can imagine it}&
\multicolumn{1}{|p{4.6cm}|}{
{\it your imagination is wrong, very wrong at that.}}                      				& -0.1\\
{\it it's saad oh yeah the snow is very beautiful}&
{\it yeah i can imagine}&
\multicolumn{1}{|p{4.6cm}|}{
{\it the weather in russia is very cool.}}                                     		& -0.7\\ \cline{1-3} 
\multicolumn{1}{|p{4.6cm}}{
walking outside with lightning and thunder sounds 10x scarier than when you 're inside}&
play in the rain&
\multicolumn{1}{p{4.6cm}|}{
haha i love listening to thunder! i would def be out there if i wasn't so sleepy} & 0.9\\ \cline{1-2}
{\it also rain(s)}&
{\it no play, rain(s)}&
\multicolumn{1}{|p{4.6cm}|}{
{\it i'm scared of the thunder}}               				    						& 0.5\\
{\it no beach or running today, stupid weather.}&
{\it lets play in the rain. ``yolo'' isnt that the vibe??}&
\multicolumn{1}{|p{4.6cm}|}{
{\it i would but its thundering to}}        													& 0.2\\
{\it then its raining...}&
{\it go outside and play in the rain lol1}&
\multicolumn{1}{|p{4.6cm}|}{
{\it if i had a buddy to go play in it with , i would . it's no fun alone}}    										         		& 0.1\\
%
{\it my only official day off this week... now what to do}&
{\it play in the rain!}&
\multicolumn{1}{|p{4.6cm}|}{
{\it yeah right haha. i wan na go swimming}}    											& 0.1\\
{\it now playing silence}&
{\it nice music i 'm playing rain}&
\multicolumn{1}{|p{4.6cm}|}{
{\it ohh i love that song}}                          									& -0.2\\ \cline{1-3}
\multicolumn{1}{|p{4.6cm}}{
yeah it is, i just dont see the point of the first 2 laps}   &
me either it would be sick if there was like 5 laps          &
\multicolumn{1}{p{4.6cm}|}{
exactly and if there were more riders so they drafted and crashed each other}    	& 0.3\\ \cline{1-2}
{\it yippee how many 711 are there}&
{\it like 5!}&
\multicolumn{1}{|p{4.6cm}|}{
{\it sweaarr i thought there were more}}        																	& 0.8\\
{\it i've got that muck off dry shower stuff to try out, hope it's muddy now.}&
{\it i think if we go for it we cud get 5 laps in an hour!}&
\multicolumn{1}{|p{4.6cm}|}{
{\it i've heard its a 30 min lap. but that was from a dh rider!}}             		& 0.6\\ 
{\it how much are they ?}&
{\it like \$5}&
\multicolumn{1}{|p{4.6cm}|}{
{\it i thought they were more then that but ok}}               										& 0.4\\
{\it igot you, wen iroll up ill pass that shit. iaint stingy.}&
{\it me either!!}&
\multicolumn{1}{|p{4.6cm}|}{
{\it they more the marrier.}}                 																		& -0.3\\
{\it i dont eat gravy on biscuits.}&
{\it me either.}&
\multicolumn{1}{|p{4.6cm}|}{
{\it well then! why were the biscuits needed?}}                  									& -0.8\\ \cline{3-3} 
\bottomrule
\end{tabular}
\caption{\camready{Sample reference sets created by our multi-reference extraction algorithm, along with the weights used in \dbleu. Triples from which additional references are extracted are in italics. Boxed sentences are in our multi-reference dev set.}} 
\label{tab:examples}
\end{table*}

To create the multi-reference \bleu dev and test sets used in this study, we adapted and extended the methodology of Sordoni et al.~\shortcite{Sordoni2015}.
From a corpus of 29M Twitter context-message-response conversational triples, we randomly extracted approximately 33K candidate triples that were then judged for conversational quality on a 5-point Likert-type scale by 3 crowdsourced annotators.
Of these, 4232 triples scored an average 4 or higher; these were randomly binned to create seed dev and test sets of 2118 triples and 2114 triples respectively. \camready{Note that the dev set is not used in the experiments of this paper, since \dbleu and IBM \bleu are metrics that do not require training. However, the dev set is released along with a test set in the dataset release accompanying this paper.}

We then sought to identify candidate triples in the 29M corpus for which both message and response are similar to the original messages and responses in these seed sets.
To this end, we employed an information retrieval algorithm with a bag-of-words BM25 similarity function~\cite{Robertson1995}, as detailed in Sordoni et al.~\shortcite{Sordoni2015}, to extract the top 15 responses for each message-response pair.
Unlike Sordoni et al.~\shortcite{Sordoni2015}, we further appended the original messages (as if parroted back).
The new triples were then scored for quality of the response in light of both context and message by 5 crowdsourced raters each on a 5-point Likert-type scale.\footnote{For this work, we sought 2 additional annotations of the seed responses for consistency with the mined responses.  As a result, scores for some seed responses slipped below our initial threshold of 4. Nonetheless, these responses were retained.}
Crucially, and again in contradistinction to Sordoni et al.~\shortcite{Sordoni2015}, we did not impose a score cutoff on these synthetic multi-reference sets.
Instead, we retained all candidate responses and scaled their scores into [$-1$, $+1$].

\camready{Table 1 presents representative multi-reference examples (from the dev set) together with their converted scores.
The context and messages associated with the supplementary mined responses are also shown for illustrative purposes to demonstrate the range of conversations from which they were taken. 
In the table, negative-weighted mined responses are semantically orthogonal to the intent of their newly assigned context and message. 
Strongly negatively weighted responses are completely out of the ballpark (``the weather in Russia is very cool'', ``well then! Why were the biscuits needed?''); others are a little more plausible, but irrelevant or possibly topic changing (``ohh I love that song''). 
Higher-valued positive-weighted mined responses are typically reasonably appropriate and relevant (even though extracted from a completely unrelated conversation), and in some cases can outscore the original response, as can be seen in the third set of examples.}

\subsection{Human Evaluation of System Outputs}
\label{sec:sysout}

Responses generated by the 7 systems used in this study on the 2114-triple test set were hand evaluated by 5 crowdsourced raters each on a 5-point Likert-type scale.  
From these 7 systems, 12 system pairs were evaluated, for a total of about pairwise 126K ratings ($12\cdot{}5\cdot{}2114$).
Here too, raters were asked to evaluate responses in terms of their relevance to both context and message.
Outputs from different systems were randomly interleaved for presentation to the raters.
We obtained human ratings on the following systems:\\[0.2cm]
%
{\bf Phrase-based MT}:
A phrase-based MT system similar to \cite{Ritter2011}, whose weights have been manually tuned. We also included four variants of that system, which we tuned with MERT~\cite{Och2003}. These variants differ in their number of features, and augment \cite{Ritter2011} with the following phrase-level features: edit distance between source and target, cosine similarity, Jaccard index and distance, length ratio, and DSSM score \cite{DSSM}.\\
{\bf RNN-based MT}: the log-probability according to the RNN model of \cite{Sordoni2015}.\\
{\bf Baseline}: a random baseline.

While \dbleu relies on human qualitative judgments, it is important to note that human judgments on multi-references (\S~\ref{sec:multiref}) and those on system outputs were collected completely independently.
We also note that the set of systems listed above specifically does not include a retrieval-based model, as this might have introduced spurious correlation between the two datasets (\S~\ref{sec:multiref} and \S~\ref{sec:sysout}).

\section{Setup}
\label{sec:systems}

We use two rank correlation coefficients---Kendall's $\tau$ and Spearman's $\rho$---to assess the level of correlation between human qualitative ratings (\S\ref{sec:sysout}) and automated metric scores. 
More formally, we compute each correlation coefficient on a series of paired observations $(m_1, q_1), \cdots, (m_N, q_N)$.
Here, $m_i$ and $q_i$ are respectively differences in automatic metric scores and qualitative ratings for two given systems $A$ and $B$ on a given subset of the test set.\footnote{\camready{For each given observation pair $(m_i,q_i)$, we randomize the order in which $A$ and $B$ are presented to the raters in order to avoid any positional bias.}}
While much prior work assesses automatic metrics for MT and other tasks \cite{Lavie2007,Hodosh2013} by computing correlations on observations consisting of single-sentence system outputs, it has been shown (e.g., \newcite{MetricsMATR}) that correlation coefficients significantly increase as observation units become larger. 
For instance, corpus-level or system-level correlations tend to be much higher than sentence-level correlations; \newcite{Graham2014} show that \bleu is competitive with more recent and advanced metrics when assessed at the system level.\footnote{We do not intend to minimize the benefit of a metric that would be competitive at the sentence-level, which would be particularly useful for detailed error analyses.
However, our main goal is to reliably evaluate generation systems on test sets of thousands of sentences, in which case any metric with good corpus-level correlation (such as \bleu , as shown in \cite{Graham2014}) would be sufficient.}

Therefore, we define our observation unit size to be $M=100$ sentences (responses),\footnote{Enumerating all possible ways of assigning sentences to observations would cause a combinatorial explosion. Instead, for all our results we sample 1K assignments and average correlations coefficients over them (using the same 1K assignments across all metrics). These assignments are done in such a way that all sentences within an observation belong to the same system pair.} unless stated otherwise.
We evaluate $q_i$ by averaging human ratings on the $M$ sentences, and $m_i$ by computing metric scores on the same set of sentences.\footnote{We refrained from using larger units, as creating larger observation units $M$ reduces the total number of units $N$. This would have caused confidence intervals to be so wide as to make this study inconclusive.}
We compare three different metrics: \bleu, \dbleu, and sentence-level \bleu (s\bleu).
The last computes sentence-level \bleu scores \cite{Nakov2012} and averages them on the $M$ sentences (akin to macro-averaging).
Finally, unless otherwise noted, all versions of \bleu use $n$-gram order up to 2 (\bleu-2), as this achieves better correlation for all metrics on this data.

\flushbottom

\section{Results}
\label{sec:results}

\begin{table}[tbp]
\small
\centering
\begin{tabular}{@{}llcc@{}}
\toprule
{\bf Metric} & {\bf refs.}  & {\bf Spearman's $\rho$} & {\bf Kendall's $\tau$}\\
\midrule
\bleu        & single       & .260 (.178, .337)       & .171 (.087, .252)\\
\bleu        & $w \ge 0.6$  & .343 (.265, .416)       & .232 (.150, .312)\\
\bleu        & all          & .318 (.239, .392)       & .212 (.129, .292)\\
\midrule
s\bleu       & single       & .265 (.183, .342)       & .175 (.091, .256)\\
s\bleu       & $w \ge 0.6$  & .330 (.252, .404)       & .222 (.140, .302)\\
s\bleu       & all          & .258 (.177, .336)       & .167 (.083, .249)\\
\midrule
\dbleu       & single       & .280 (.199,.357)        & .187 (.103, .268)\\
\dbleu       & $w \ge 0.6$  & .405 (.331,.474)        & .281 (.200, .357)\\
\dbleu       & all          & {\bf .484} (.415,.546)  & {\bf .342} (.265, .415)\\
\bottomrule
\end{tabular}
\caption{Human correlations for IBM \bleu, sentence-level \bleu, and \dbleu with 95\% confidence intervals. This compares 3 types of references: single only, high scoring references ($w\ge0.6$), and all references.}
\label{tab:results}
\end{table}

\begin{figure}[tbp]
\centering
\includegraphics[trim=3 30 10 37, clip,scale=0.65]{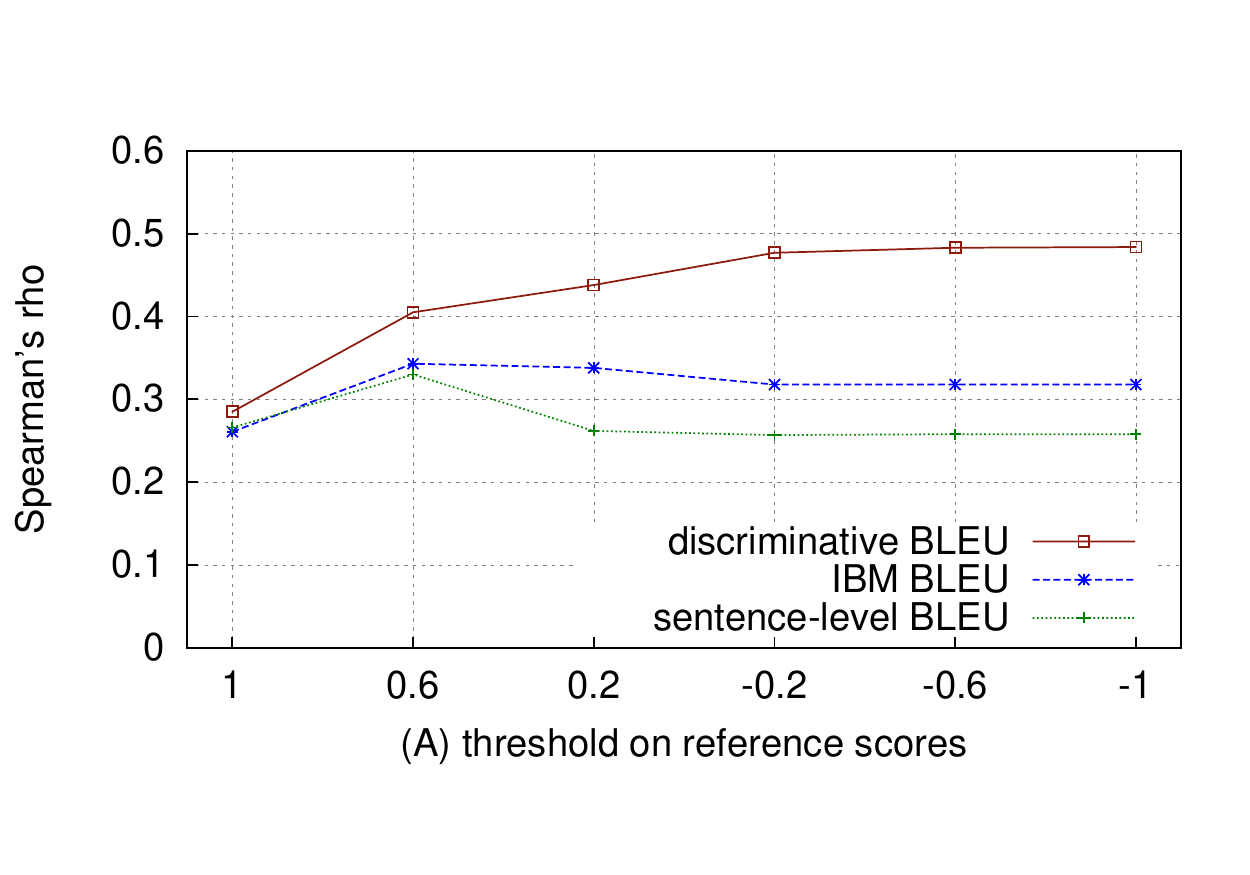}
\includegraphics[trim=3 30 10 37, clip,scale=0.65]{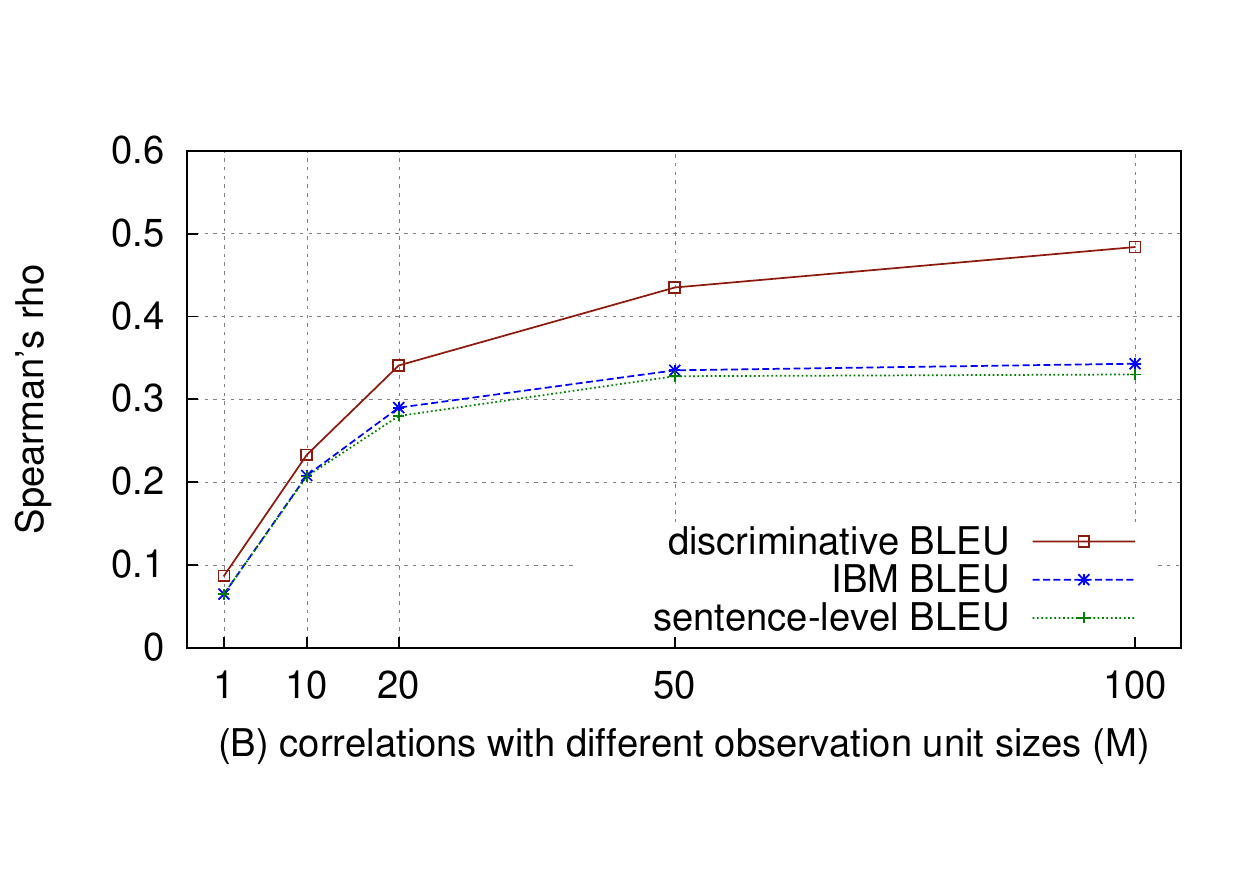}
\includegraphics[trim=3 30 10 37, clip,scale=0.65]{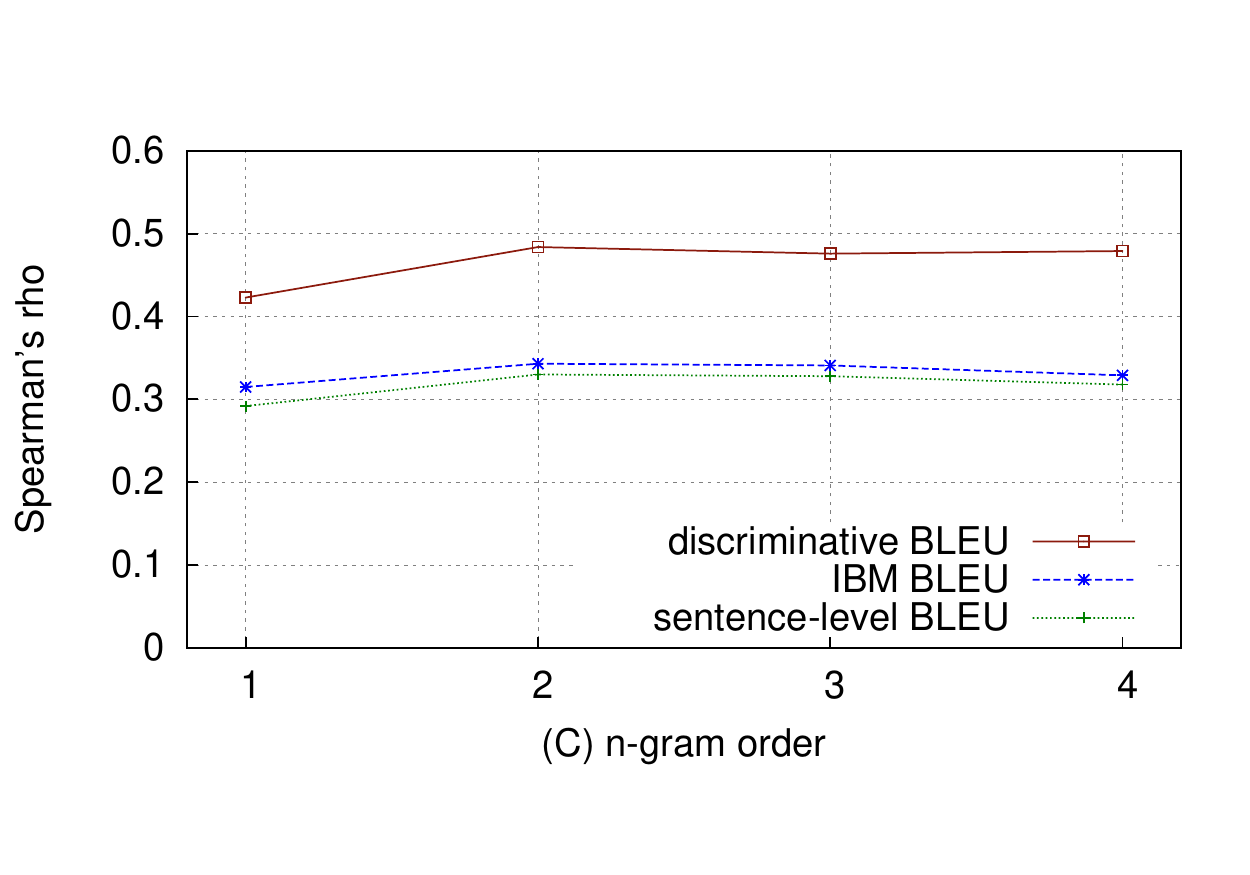}
\caption{A comparison of \bleu, sentence-level \bleu, and \dbleu along three dimensions: (A) decreasing the threshold on reference scores $w_{i,j}$; (B) increasing the unit size for the correlation study from a single sentence ($M$=1) to a size of 100; (C) going from \bleu-1 to \bleu-4 for the different versions of \bleu.}
\label{tab:plots}
\end{figure}


The main results of our study are shown in Table~\ref{tab:results}. 
\dbleu achieves better correlation with human than \bleu, when comparing the best configuration of each metric.\footnote{This is also the case on single reference. While \dbleu and \bleu would have the same correlation if original references all had the same score of 1, it is not unusual for original references to get ratings below 1.}
In the case of Spearman's $\rho$, the confidence intervals of \bleu $(.265, .416)$ and \dbleu $(.415,.546)$ barely overlap, while interval overlap is more significant in the case of Kendall's $\tau$.
Correlation coefficients degrade for \bleu as we go from $w \ge 0.6$ to using all references.
This is expected, since \bleu treats all references as equal and has no way of discriminating between them.
On the other hand, correlation coefficients increase for \dbleu after adding lower scoring references.
It is also worth noticing that \bleu and s\bleu obtain roughly comparable correlation coefficients.
This may come as a surprise, because it has been suggested elsewhere that s\bleu has much worse correlation than \bleu computed at the corpus level~\cite{MetricsMATR}.
We surmise that (at least for this task and data) the differences in correlations between \bleu and s\bleu observed in prior work may be less the result of a difference between micro- and macro-averaging than they are the effect of different observation unit sizes (as discussed in \S\ref{sec:systems}).

Finally, Figure~\ref{tab:plots} shows how Spearman's $\rho$ is affected along three dimensions of study.
In particular, we see that \dbleu actually benefits from the references with negative ratings. 
While the improvement is not pronounced, we note that most references have positive ratings.
Negatively-weighted references could have a greater effect if, for example, randomly extracted responses had also been annotated.

\section{Conclusions}
\label{sec:conclusion}
\dbleu correlates well with human quality judgments of generated conversational responses, outperforming both IBM \bleu and sentence-level \bleu in this task and demonstrating that it can serve as a plausible intrinsic metric for system development.\footnote{\camready{An implementation of \dbleu, multi-reference dev and test sets, and human rated outputs are available at:\\ \url{http://research.microsoft.com/convo}}} 
An upfront cost is paid for human evaluation of the reference set, but following that, the need for further human evaluation can be minimized during system development. 
\dbleu may help other tasks that use multiple references for intrinsic evaluation, including image-to-text, sentence compression, and paraphrase generation, and even statistical machine translation.
Evaluation of \dbleu in these tasks awaits future work.

\section*{Acknowledgments}
We thank the anonymous reviewers, Jian-Yun Nie, and Alan Ritter for their helpful comments and suggestions.


\bibliographystyle{acl}
\bibliography{acl15-eval}

\end{document}